\documentclass[runningheads]{llncs}
\usepackage[T1]{fontenc}
\usepackage{graphicx}%
\usepackage{multirow}%
\usepackage{amsmath,amssymb,amsfonts}%
\usepackage{mathrsfs}%
\usepackage[title]{appendix}%
\usepackage{xcolor}%
\usepackage{textcomp}%
\usepackage{manyfoot}%
\usepackage{booktabs}%
\usepackage{algorithm}%
\usepackage{algorithmicx}%
\usepackage{algpseudocode}%
\usepackage{listings}%
\usepackage{xcolor}
\begin{document}
\title{Symbolic-AI-Fusion Deep Learning (SAIF-DL): Encoding Knowledge into Training with Answer Set Programming Loss Penalties by a Novel Loss Function Approach}

\titlerunning{Symbolic-AI-Fusion Deep Learning: Encoding Knowledge into Training}
%
\author{Fadi Al Machot\orcidID{0000-0002-1239-9261} \and Martin Thomas Horsch\orcidID{0000-0002-9464-6739} \and 
Habib Ullah\orcidID{0000-0002-2434-0849}} 
\authorrunning{F.~Al Machot et al.}
%
\institute{Department of Data Science, Faculty of Science and Technology, Norwegian University of Life Sciences, P.O.~Box 5003, 1432 \AA{}s, Norway \\
\email{\{fadi.al.machot, martin.thomas.horsch, habib.ullah\}@nmbu.no}}
\maketitle              
%

\begin{abstract}
This paper presents a hybrid methodology that enhances the training process of deep learning (DL) models by embedding domain expert knowledge using ontologies and answer set programming (ASP). By integrating these symbolic AI methods, we encode domain-specific constraints, rules, and logical reasoning directly into the model's learning process, thereby improving both performance and trustworthiness. The proposed approach is flexible and applicable to both regression and classification tasks, demonstrating generalizability across various fields such as healthcare, autonomous systems, engineering, and battery manufacturing applications. Unlike other state-of-the-art methods, the strength of our approach lies in its scalability across different domains. The design allows for the automation of the loss function by simply updating the ASP rules, making the system highly scalable and user-friendly. This facilitates seamless adaptation to new domains without significant redesign, offering a practical solution for integrating expert knowledge into DL models in industrial settings such as battery manufacturing.
\end{abstract}

\section{Introduction}
Machine learning models, particularly deep learning~(DL), have demonstrated remarkable success across a variety of fields, including image recognition, natural language processing, and predictive analytics~\cite{lecun2015deep}. However, in domains where rules, constraints, and logical reasoning are critical, such as healthcare, autonomous systems, engineering, and finance, these models can face significant limitations. Traditional DL models are primarily data-driven and often function as black boxes, lacking the ability to incorporate explicit domain knowledge or reasoning capabilities~\cite{darwiche2018human}.
One of the key challenges is that DL models may overlook domain-specific knowledge that is not easily captured within the dataset, leading to suboptimal or even incorrect predictions. For instance, a DL model predicting medication dosages could fail to recognize contraindications between medications unless domain knowledge is introduced. This limitation underscores the need for integrating symbolic reasoning and domain knowledge into the learning process to enhance model performance and trustworthiness.

To address this challenge, we present a hybrid methodology that integrates deep learning with domain expert knowledge using ontologies~\cite{baader2017dl,gruber1993translation} and answer set programming~(ASP)~\cite{gelfond1988stable}. Our approach involves embedding domain-specific constraints and logical rules directly into the loss function of the DL model. By doing so, we create a loss function that balances predictive accuracy (data-driven learning) with rule adherence (knowledge-driven learning). This integration ensures that the model not only learns from the data but also adheres to domain-specific constraints, enhancing its suitability for high-stakes applications.

By combining data-driven learning with symbolic reasoning, our methodology produces models that are both accurate and aligned with domain knowledge. This integration enhances explainability, as the logical rules provide insights into the model's decision-making process, thereby ensuring higher levels of trustworthiness and interpretability~\cite{arrieta2020explainable}.
Furthermore, unlike other state-of-the-art methods that may require extensive redesign when applied to different domains, our approach offers scalability and ease of adaptation. By simply updating the ASP rules within the ontology, the same pipeline can be applied across various fields without significant modifications. This feature is particularly advantageous in industrial settings like battery manufacturing, where rapid adaptation to new processes or regulations is essential \cite{cho2022physics}.

\section{Related Work}

Purely data-driven modelling using neural networks is agnostic with respect to the laws and rules underlying the data and, by design, primarily meant for interpolation, with a comparably restricted potential for transfer learning. As a consequence, purely data-driven modelling requires a large volume of input data in order to sufficiently cover the domain within which interpolation is to be facilitated. As another consequence, it may fail to correctly reproduce facts or internal consistency relationships that appear trivial to the end user, creating embarrassing counterexamples. 

Neural-symbolic approaches seek to bridge these limitations by combining the adaptive learning capabilities of neural networks with the logical, rule-based reasoning of symbolic AI. This integration allows AI systems to tackle a broader range of tasks, encompassing both low-level perception (e.g., image and speech recognition) and high-level reasoning (e.g., inference and abstract concept manipulation)~\cite{bhuyan2024neuro}. For instance, neuro-symbolic systems have shown promising results in visual question answering~(VQA), where systems must interpret visual data and respond to natural language queries, blending perceptual and symbolic reasoning to achieve accurate results~\cite{dinu2024symbolicai}.  Additionaly, such approaches perform well with limited training data due to the guidance provided by symbolic reasoners~\cite{machot2023bridging}. 

Various symbolic AI models have been developed, each with unique strengths in knowledge representation and reasoning. For instance, logic programming using Prolog~\cite{kowalski1974predicate} supports problem-solving and formal reasoning by allowing the definition of facts, rules, and queries. Similarly, ASP emphasizes nonmonotonic reasoning for dynamic problem-solving in areas such as planning and constraint satisfaction~\cite{machot2023bridging}.
One significant area of development within neural-symbolic AI is concept learning, which merges the pattern recognition strengths of neural networks with the abstract representation capabilities of symbolic reasoning. This combination is particularly beneficial for compositional reasoning tasks, where the goal is to understand relationships between different concepts. Notable systems, such as the Neuro-Symbolic Concept Learner~(NS-CL), effectively integrate scene understanding with symbolic reasoning to perform tasks like concept learning and relational reasoning, thereby enabling AI models to generalize knowledge to new contexts~\cite{mao2019neuro}.
A significant advancement in this domain is the advent of deep probabilistic programming languages~(DPPLs)~\cite{manhaeve2021neural}, which merge neural predicates into a probabilistic logic programming framework. This fusion allows for the flexibility of deep learning while incorporating the structured reasoning strengths of probabilistic inference, facilitating both symbolic and subsymbolic reasoning. NeurASP~\cite{yang2023neurasp} enhances traditional Answer Set Programs by utilizing neural networks to manage uncertainties in symbolic reasoning, interpreting neural outputs as probabilistic distributions for atomic facts. This method has been applied in areas such as probabilistic knowledge representation and decision-making systems.
In addition, physics-informed neural networks~(PINNs)~\cite{cai2021physics} are a class of neural networks that integrate fundamental physical laws into the learning process of deep learning models. Instead of relying solely on data to train the model, PINNs incorporate physical principles described by partial differential equations~(PDEs) or other mathematical formulations directly into the network's architecture or loss function. This approach ensures that the model's predictions are consistent with known physics, enhancing both accuracy and generalizability.

In contrast, our methodology, symbolic-AI-fusion deep learning (SAIF-DL), allows for the seamless integration of new domain knowledge by simply updating the ASP rules within the ontology. This modularity means that the same underlying neural network architecture can be reused across different domains with minimal adjustments. The ease with which domain knowledge can be updated or expanded without altering the core model enhances the scalability and user-friendliness of our approach.
\section{Overall Methodology}

The hybrid methodology presented in this work involves the integration of domain expert knowledge using ontologies and Answer Set Programming (ASP) into the training process of deep learning models. This is accomplished by encoding domain-specific constraints and rules into the loss function of the model, which guides the training process.

The first step in this approach involves creating or enriching an ontology that captures the key concepts, relationships, and constraints relevant to the domain. Ontologies are enhanced with domain expert input to ensure they accurately reflect the problem space. This knowledge is then encoded into ASP rules, which formalize the constraints as logical expressions. The DL model is trained with a customized loss function that not only minimizes prediction error but also applies a penalty when ASP rules are violated.

The ASP rules are used during the training process to evaluate the model’s predictions, ensuring that they adhere to domain-specific constraints. If the predictions violate any rules, a penalty is added to the loss function, which helps guide the model to produce more compliant outputs in future iterations.

This iterative training process continues until the model learns to balance between fitting the data and adhering to domain knowledge, resulting in a model that is both accurate and reliable.

\subsection{Pipeline Overview}

\begin{figure}[h!]
    \centering
    \includegraphics[width=\textwidth]{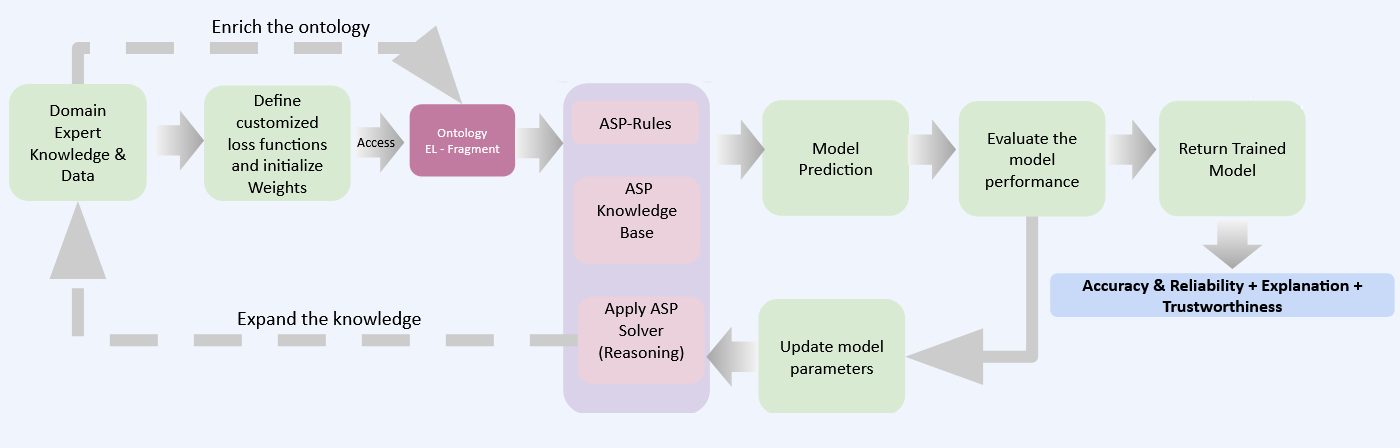}
    \caption{An overview of the hybrid methodology integrating domain expert knowledge through ontologies and Answer Set Programming (ASP) into the deep learning training process. The pipeline illustrates how domain knowledge is used to enrich the ontology, apply ASP rules, and provide penalties in the loss function to ensure compliance with domain-specific constraints. This iterative process leads to a model that balances accuracy with reliability, explainability, and trustworthiness.}
    \label{fig:method_sketch}
\end{figure}

The pipeline for this hybrid methodology is illustrated in Figure \ref{fig:method_sketch}. The figure provides a step-by-step visual representation of the key components and processes involved in the methodology.

As seen in Figure \ref{fig:method_sketch}, the pipeline begins with domain expert knowledge and data, which are used to define an ontology. This ontology is continuously enriched and accessed throughout the training process. Customized loss functions are defined, which are influenced by the expert knowledge captured in the ontology. A fragment of the ontology is translated into ASP rules, which serve as a formal mechanism for reasoning over domain-specific constraints.

The ASP rules and the knowledge base are fed into an ASP solver, which checks whether the model’s predictions violate any of the encoded rules. If violations are detected, they are reflected in an ASP penalty, which is incorporated into the total loss function. This feedback is then used to update the model parameters, guiding the training process to ensure compliance with both data-driven predictions and domain knowledge.

After training, the model is evaluated based on its performance in prediction tasks as well as its adherence to the domain rules. The final result is a trained model that demonstrates a balance between accuracy, reliability, explainability, and trustworthiness, making it well-suited for high-stakes applications.

\section{Designing the Loss Function for Hybrid Learning}

The loss function in this hybrid approach is designed to incorporate domain knowledge from ASP into the learning process, alongside the traditional data-driven loss. The base loss function, $\mathcal{L}(y_{\text{true}}, y_{\text{pred}})$, is chosen based on the nature of the task at hand. For classification problems, cross-entropy loss is typically used to measure the divergence between predicted probabilities and true labels. For regression tasks, mean squared error (MSE) or mean absolute error (MAE) is commonly employed to measure the difference between predicted values and ground truth \cite{machot2023bridging}.

To ensure that the model adheres to domain knowledge, an ASP-based penalty term is incorporated into the loss function. The total loss function is expressed as follows:

\begin{equation}
\text{Total Loss} = \mathcal{L}(y_{\text{true}}, y_{\text{pred}}) + \lambda \cdot \text{ASP Penalty}
\end{equation}

In this equation, $\lambda$ is a weighting factor that controls the importance of the ASP penalty in the learning process. The ASP penalty term increases if the model’s predictions violate ASP rules, reflecting the degree to which the model’s predictions deviate from domain knowledge. By tuning $\lambda$, we can adjust the balance between the model’s ability to learn from the data and its adherence to domain-specific constraints.

\subsection{Designing Differentiable Penalty Functions}

The key to making the ASP penalty differentiable is to represent rule violations using smooth, continuous functions. Instead of binary indicators of rule compliance, we use differentiable functions that quantify the degree of violation.

For each ASP rule, we define a penalty function $P_i(y_{\text{pred}})$ that measures how much the model's prediction violates the rule. The ASP penalty is then the sum of these individual penalties:

\begin{equation}
\text{ASP Penalty} = \sum_{i=1}^{N} \gamma_i \cdot P_i(y_{\text{pred}})
\end{equation}

where:

\begin{itemize}
    \item $N$ is the total number of ASP rules.
    \item $\gamma_i$ is the weight assigned to the $i$-th rule, reflecting its importance.
    \item $P_i(y_{\text{pred}})$ is a differentiable penalty function for the $i$-th rule.
\end{itemize}

\subsubsection{Example of a Differentiable Penalty Function}

Consider the example in the context of battery manufacturing:

\textbf{Domain Rule}:

\begin{quote}
``The charging voltage must not exceed 4.2V to prevent battery degradation.''
\end{quote}

We want to design a differentiable penalty function that penalizes predictions where the voltage exceeds 4.2V. A common approach is to use the ReLU (Rectified Linear Unit) function or a smooth approximation to model the violation.

The penalty function $P_{\text{voltage}}(V_{\text{predicted}})$ can be defined as:

\begin{equation}
P_{\text{voltage}}(V_{\text{predicted}}) = \text{ReLU}(V_{\text{predicted}} - V_{\text{max}})
\end{equation}

or, to ensure smoothness, we can use the Softplus function:

\begin{equation}
P_{\text{voltage}}(V_{\text{predicted}}) = \frac{1}{k} \ln\left(1 + e^{k (V_{\text{predicted}} - V_{\text{max}})}\right)
\end{equation}

where:

\begin{itemize}
    \item $V_{\text{max}} = 4.2$V is the maximum allowable voltage.
    \item $k$ is a hyperparameter that controls the sharpness of the penalty function.
\end{itemize}

The Softplus function is a smooth, differentiable approximation of the ReLU function and ensures that the penalty increases smoothly as the predicted voltage exceeds $V_{\text{max}}$.

\subsection{Adjusting the Weighting Factor $\lambda$}

The weighting factor $\lambda$ balances the influence of data fitting and rule adherence:

\begin{itemize}
    \item A larger $\lambda$ emphasizes compliance with domain rules, which is important in safety-critical applications.
    \item A smaller $\lambda$ allows the model to prioritize data fitting, which may be desirable when the data contains patterns not fully captured by the rules.
\end{itemize}

Selecting an appropriate $\lambda$ involves experimentation and validation to achieve the desired balance.

By incorporating these differentiable penalty functions into the loss function, the model learns to make predictions that optimize production parameters while complying with essential manufacturing guidelines.

\subsection{Overall Algorithm}

Algorithm \ref{alg:hybrid_training} summarizes the training process with differentiable ASP penalty integration.

\begin{algorithm}[h]
\caption{Hybrid Training with Differentiable ASP Penalty}
\label{alg:hybrid_training}
\begin{algorithmic}[1]
\Require Training data $(X, y_{\text{true}})$, initial model parameters $\theta$, differentiable penalty functions $P_i(y_{\text{pred}})$, weights $\gamma_i$, ASP penalty weight $\lambda$
\While{not converged}
    \State \textbf{Forward Pass}:
    \State Compute predictions $y_{\text{pred}} = f_{\theta}(X)$
    \State \textbf{ASP Penalty Computation}:
    \State Compute $P_i(y_{\text{pred}})$ for all rules
    \State Compute total ASP penalty $\text{ASP Penalty} = \sum_{i=1}^{N} \gamma_i \cdot P_i(y_{\text{pred}})$
    \State \textbf{Loss Computation}:
    \State Compute base loss $\mathcal{L}(y_{\text{true}}, y_{\text{pred}})$
    \State Compute total loss $\text{Total Loss} = \mathcal{L} + \lambda \cdot \text{ASP Penalty}$
    \State \textbf{Backward Pass}:
    \State Compute gradients $\nabla_{\theta} \text{Total Loss}$
    \State Update model parameters $\theta \leftarrow \theta - \eta \nabla_{\theta} \text{Total Loss}$
\EndWhile
\end{algorithmic}
\end{algorithm}

\section{Proof of Concept Experiment}

In this experiment, we integrate ASP constraints into a neural network training process to enforce domain-specific knowledge. A synthetic dataset with 1000 instances was generated, where each instance consists of two features $x_1$ and $x_2$ randomly sampled from a uniform distribution between 0 and 1. The target label is $1$ if $x_1 + x_2 > 1$, and $0$ otherwise.

A simple feedforward neural network was implemented with an input layer of two nodes, one hidden layer of 10 neurons with ReLU activation, and an output layer for binary classification. The network was trained using the Adam optimizer and cross-entropy loss over 20 epochs. In addition to training without constraints, the ASP-based penalty was incorporated to enforce domain knowledge that $x_1 > 0.8$ should predict the class label $1$. The ASP penalty was computed using Clingo~\cite{gebser2019clingo}, where logical rules encoded the constraint and penalized violations during training.

The model was trained in two configurations: First, using only cross-entropy loss, and second, with an additional ASP penalty. Accuracy and domain satisfaction were measured in both cases. Results showed that incorporating ASP penalties improved domain satisfaction from 0.78 to 0.95, ensuring the model’s predictions adhered more closely to the domain rule. Accuracy remained competitive, reaching 92.7\% with the ASP penalty compared to 89.3\% without it. This proof of concept demonstrates the potential of integrating symbolic reasoning into neural network training to enforce domain-specific constraints.

\begin{figure}[h!]
    \centering
    \includegraphics[width=\textwidth]{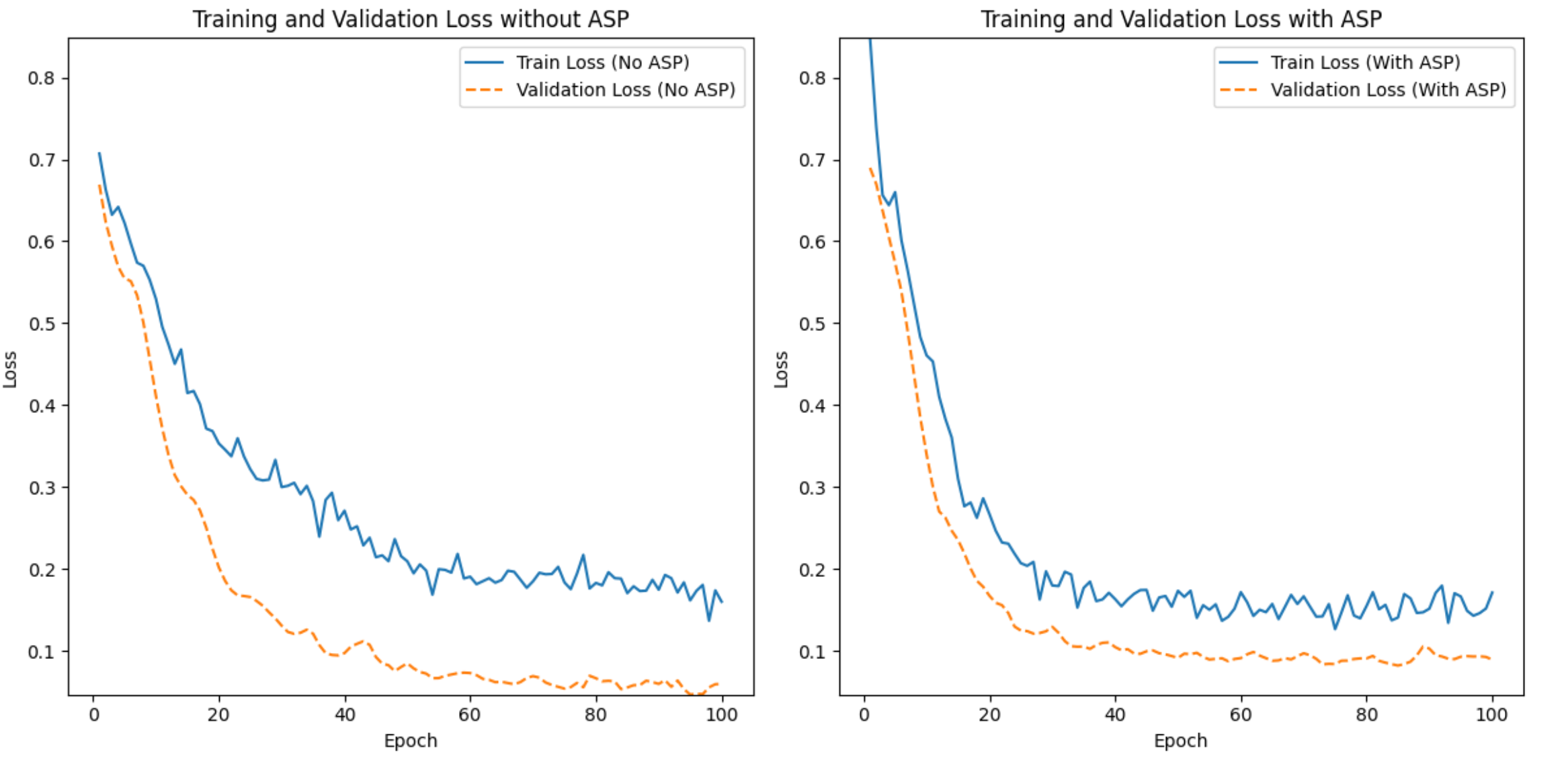}
    \caption{Comparison of Training and Validation Loss Curves for Models With and Without ASP Penalty}
    \label{fig:res}
\end{figure}

Figure \ref{fig:res} illustrates the training and validation loss curves for models trained with and without the ASP penalty. In the left plot, without ASP, the training loss decreases steadily, but the validation loss flattens earlier, suggesting potential overfitting and moderate generalization. In contrast, the right plot, with ASP, shows closely aligned training and validation losses, highlighting reduced overfitting and enhanced generalization.

\section{Application in Various Domains}

The methodology outlined here is applicable across a range of domains where domain-specific knowledge plays a critical role. In healthcare, for example, the model can integrate medical ontologies and ASP rules to ensure that predictions adhere to clinical guidelines and avoid harmful drug interactions.
Specifically, discussing what constitutes reliability, and therefore admissibility, of an AI system in the medical field, Dur\'an and Jongsma~\cite{duran2021medical} refer to ``verification and validation methods, robustness analysis, a history of (un)successful implementations, and expert knowledge'' as factors of computational reliabilism~\cite{duran2018grounds}. SAIF-DL facilitates reliable system design across multiple of these factors: verification, validation, and integration of expert knowledge.
For autonomous systems, ASP can encode safety rules and regulations, such as speed limits or obstacle avoidance protocols, ensuring that the system operates safely even in complex and dynamic environments.
In finance, regulatory compliance is critical, and models must adhere to strict legal and ethical guidelines. ASP rules can be used to encode these constraints, ensuring that the model’s predictions comply with legal requirements, such as those related to credit scoring or anti-money laundering regulations.

The adaptability of this hybrid methodology makes it suitable for any domain where domain knowledge, safety, compliance, or trustworthiness are critical to the success of AI systems.

\section{Conclusion}

This paper has presented a novel hybrid methodology that integrates deep learning with domain expert knowledge using ontologies and answer set programming. By embedding domain-specific constraints and logical rules directly into the loss function of deep learning models, we have developed a flexible and scalable approach that balances predictive accuracy with rule adherence. This integration ensures that models not only learn from data but also comply with critical domain knowledge, enhancing their suitability for high-stakes applications such as battery manufacturing.

Our approach leverages the expressive power of ontologies and ASP to represent a broader range of domain knowledge, including logical constraints and expert rules that are not easily captured by differential equations. This flexibility allows for seamless adaptation to various domains by simply updating the ASP rules, without the need for significant modifications to the neural network architecture.

By designing differentiable penalty functions for the ASP rules, we have ensured that the total loss function remains compatible with gradient-based optimization methods. This enables effective training of the hybrid model, allowing it to learn smoothly from both data and domain knowledge. The proposed methodology enhances the model's reliability, trustworthiness, and explainability, making it a robust solution for complex, knowledge-intensive applications.

The application to battery manufacturing demonstrates the practical utility of our approach in an industrial setting where compliance with domain-specific rules is critical. The ability to incorporate constraints such as temperature limits, mixing ratios, and process timings directly into the model's training process ensures that predictions not only optimize performance but also adhere to essential manufacturing guidelines.

Future work may explore the extension of this methodology to other domains and the incorporation of additional forms of domain knowledge. Investigating strategies for automatic tuning of the weighting factors and penalty functions could further enhance the model's performance. Additionally, integrating uncertainty quantification methods could provide deeper insights into the model's predictions, further increasing confidence in its applicability to real-world scenarios.

\begin{credits}
\subsubsection{\ackname} Funding is acknowledged from the European Union's Horizon Europe research and innovation programme under grant agreements no.~101137725 (BatCAT) and 101138510 (DigiPass CSA).

\subsubsection{\discintname}
The authors have submitted a grant proposal in which the presented architecture plays a central role.
\end{credits}
%
%
%
\bibliographystyle{splncs04}
\bibliography{ref}

\end{document}